\documentclass[sigconf]{aamas}  % do not change this line!
\renewcommand\footnotetextcopyrightpermission[1]{} % removes footnote with conference information in first column
\usepackage{booktabs}
\usepackage{times}
\usepackage{helvet}
\usepackage{courier}
\usepackage{framed}
\usepackage{epsfig}
\usepackage{amsmath}
\usepackage{graphicx}
\usepackage{amsfonts}
\usepackage{wrapfig}
\usepackage{multirow}
\usepackage{framed}
\usepackage{algorithm}
\usepackage{float}
\usepackage{subfigure}
\usepackage{amsfonts}
\usepackage{amssymb}
\usepackage{algpseudocode}
\usepackage{booktabs,arydshln}
\usepackage{amsthm}
\usepackage[font=small,labelfont=bf]{caption}

\newcommand{\pii}{\pi_{i}}
\newcommand{\pij}{\pi_{j}}

\newtheorem{defn}{Definition}
\newtheorem{eg}{Example}

\newcommand{\algrule}[1][.2pt]{\par\vskip.5\baselineskip\hrule height #1\par\vskip.5\baselineskip}

\setcopyright{ifaamas}  % do not change this line!
\acmDOI{}  % do not change this line!
\acmISBN{}  % do not change this line!
\acmConference[AAMAS'18]{Proc.\@ of the 17th International Conference on Autonomous Agents and Multiagent Systems (AAMAS 2018), M.~Dastani, G.~Sukthankar, E.~Andre, S.~Koenig (eds.)}{July 2018}{Stockholm, Sweden}  % do not change this line!
\acmYear{2018}  % do not change this line!
\copyrightyear{2018}  % do not change this line!
\acmPrice{}  % do not change this line!

\begin{document}

\title{Preference-Guided Planning: An Active Elicitation Approach}  % put your title here!
%\titlenote{Produces the permission block, and copyright information}

% AAMAS: as appropriate, uncomment one subtitle line; check the CFP
%\subtitle{Extended Abstract}
%\subtitle{Industrial Applications Track}
%\subtitle{Socially Interactive Agents Track}
%\subtitle{Blue Sky Ideas Track}
%\subtitle{Robotics Track}
%\subtitle{JAAMAS Track}
%\subtitle{Doctoral Mentoring Program}

%\subtitlenote{The full version of the author's guide is available as \texttt{acmart.pdf} document}

% AAMAS: submissions are anonymous for most tracks
%\author{Paper \#160}  % put your paper number here!

\author{Mayukh Das}
%\authornote{Dr.~Trovato insisted his name be first.}
\affiliation{%
  \institution{University of Texas, Dallas}
%  \streetaddress{P.O. Box 1212}
}
\email{mayukh.das1@utdallas.edu}
\author{Phillip Odom}
%\authornote{The secretary disavows any knowledge of this author's actions.}
\affiliation{%
  \institution{Indiana University, Bloomington}
%  \streetaddress{P.O. Box 1212}
}
\email{phodom@indiana.edu}
\author{Md. Rakibul Islam}
%\authornote{This author is the
%  one who did all the really hard work.}
\affiliation{%
  \institution{Washington State University}
%  \streetaddress{1 Th{\o}rv{\"a}ld Circle}
%  \country{Iceland}
}
\email{mislam@eecs.wsu.edu}
\author{Janardhan Rao (Jana) Doppa}
\affiliation{%
  \institution{Washington State University}
}
\email{jana@eecs.wsu.edu}

\author{Dan Roth} 
\affiliation{%
 \institution{University of Pennsylvania}
% \country{India}
}
\email{danroth@seas.upenn.edu}

\author{Sriraam Natarajan}
\affiliation{%
  \institution{University of Texas, Dallas}
%  \streetaddress{30 Shuangqing Rd}
%  \country{China}
}
\email{sriraam.natarajan@utdallas.edu}

\begin{abstract}  % put your abstract here!
Planning with preferences has been employed extensively to quickly generate high-quality plans. 
However, it may be difficult for the human expert to supply this information without knowledge of the reasoning employed by the planner and the distribution of planning problems. We consider the problem of actively eliciting preferences from a human expert during the planning process. Specifically, we study this problem in the context of the Hierarchical Task Network (HTN) planning framework as it allows easy interaction with the human.
Our experimental results on several diverse planning domains show that the preferences gathered using the proposed approach improve the quality and speed of the planner, while reducing the burden on the human expert.
\end{abstract}

% AAMAS: the ACM CCS are not needed within AAMAS papers
%%
%% The code below should be generated by the tool at
%% http://dl.acm.org/ccs.cfm
%\begin{CCSXML}
%<ccs2012>
%  <concept_id>10010520.10010553.10010562</concept_id>
%  <concept_desc>Computer systems organization~Embedded systems</concept_desc>
%  <concept_significance>500</concept_significance>
%  <concept_id>10010520.10010575.10010755</concept_id>
%  <concept_desc>Computer systems organization~Redundancy</concept_desc>
%  <concept_significance>300</concept_significance>
%  <concept_id>10010520.10010553.10010554</concept_id>
%  <concept_desc>Computer systems organization~Robotics</concept_desc>
%  <concept_significance>100</concept_significance>
%  <concept_id>10003033.10003083.10003095</concept_id>
%  <concept_desc>Networks~Network reliability</concept_desc>
%  <concept_significance>100</concept_significance>
%</ccs2012>  
%\end{CCSXML}
%
%\ccsdesc[500]{Computer systems organization~Embedded systems}

  % put your semicolon-separated keywords here!

\maketitle

%%%%%%%%%%%%%%%%%%%%%%%%%%%%%%%%%%%%%%%%%%%%%%%%%%%%%%%%%%%%%%%%%%%%%%%%%%%%%%%%%%%%%%%%%%%%%%%%%%%%%%%%%
%% start of main body of paper

%\input{samplebody-conf}
% --------------INTRODUCTION --------------------------------
\section{Introduction}

Planning under uncertainty has exploited human (domain) expertise  in several different directions~\cite{tan1994specification,dean1995planning,boutilier1995planning,myers1996advisable,huang1999control,allen2002human,brafman2005planning,sohrabi2008planning}. This contrasts with traditional learning techniques that require large amount of labeled data and treat the human as a ``mere labeler". One key research thrust in this direction is that of specifying preferences as advice to the planner in order to reduce the search over the space of plans. While successful, most of the preference specification approaches require that the human input be provided in advance before planning commences. There are at least two main issues with this approach: (1) the human sometimes provides the most ``obvious" advice that can be potentially inferred by calculating the uncertainty in the plan space, and (2) the planner may not reach the plan space where the preferences apply. 
%(1) The human sometimes provides the most ``obvious" advice that can be potentially inferred, (2) The upfront advice may not benefit the planner because the uncertainty in the plan may be drastically different from that of the human input. 

We propose a framework in which the planner actively solicits preferences as needed. More specifically, our proposed planning approach computes the uncertainty in the plan explicitly and then queries the human expert for advice as needed. This approach not only removes the burden on the human expert to provide all the advice upfront but also allows the learning algorithm to focus on the most uncertain regions of the plan space and query accordingly. Thus, it avoids human effort on trivial regions and improves the relevance of the preferences.
%instead focus on the most uncertain regions of the state space.
%Thus it avoids the humans from providing advice about trivial/most obvious regions of the plan space and instead focus on the harder part of the search from the planner's perspective.

We present an algorithm for active preference elicitation in planning %and we call our work as 
called the {\em preference-guided planner} (\textsc{PGPlanner}) to denote that the agent treats the human advice as {\em soft preferences} and solicits these preferences as needed. %{\em To the best of our knowledge, this is the first work in this direction for planning}. 
We consider a Hierarchical Task Network (HTN) planner for this task as it allows for seamless natural interaction with humans who solve problems by decomposing them into smaller problems. 
Hence, HTN planners can facilitate humans in providing knowledge at varying levels of generality. We evaluate our algorithm on several standard domains and a novel blocksworld domain where we compare against several baselines. Our results show that this collaborative approach allows for more efficient and effective problem solving compared to the standard planning as well as providing all the preferences in advance.

\noindent  {\bf Contributions:} Our key contributions include: (1) We introduce active preference elicitation for HTN planning; (2) Our framework treats the human input as soft preferences and allows to trade-off between potentially a sub-optimal expert and a complex plan space; and (3) We evaluate our algorithm on several tasks and demonstrate its efficacy against other baselines.

% --------------RELATED WORK --------------------------------
\section{Background and related work}

%We first present an overview of planning. Then, we describe related work on preference elicitation for both planning as well as active advice-seeking for decision-making.

%\paragraph{Planning:} 
%Planning problems and automated planners are often characterized using state space models. However, the problems are presented to planners in the form of compact description in a suitable language (e.g., PDDL). A planning task can be seen as a 5-tuple $\left\langle S, A, I, G, T\right\rangle$ where: $S$ is a (finite set of states; $A$ is a finite set of actions, where $A(s) \subseteq A$ corresponds to actions that are applicable at state $s \in S$; $I \subseteq S$ is the set of possible initial states; $G \subseteq S$ is the set of possible goal states; $T: S \times A \mapsto S$ is the transition function, where $T(s,a)$ returns the next state for deterministic domains, and returns the probability distribution over the next states for stochastic domains. Additionally, in partially observable domains, some information of the state is hidden. 

Given an initial state and a goal specification as a specific instance of a planning task, automated planners produce a sequence of actions (aka. plan) to satisfy the goal specification. The most basic form of automated planning is computationally hard (specifically, PSPACE-complete \cite{bylander1991complexity}). However, real-world applications require fast planners to satisfy time constrains. Consequently, there is a large body of work to address this challenge \cite{ghallab2004automated}. Some representative approaches include reduction to SAT solving \cite{kautz1992satplan,kautz1996pushing,blum1997fast}; forward state space search with human-designed heuristics \cite{hoffmann2001ff,yoon2007ff}, learned heuristics from solved planning problems \cite{yoon2008learning,xu2009learning,xu2010iterative}; planning with human-written control knowledge \cite{erol1994htn,bacchus2000using}; and solving probabilistic planning via reduction to deterministic planning \cite{yoon2008probabilistic}. Learning-based and knowledge-based approaches have seen great success, but they assume a fixed knowledge representation and can be brittle. 

Our \textsc{PGPlanner} relates closely to
%Our approach shares systemic connections to 
mixed-initiative planning~\cite{ferguson1996trains,ai2004mapgen,talamadupula2013architectural}, which interleaves planning by the expert with automated planning. However, they are conceptually different in the expert's role and intervention in the planning process. Mixed-initiative planning is a negotiation between an agent and human on mutable goals/sub-goals, and partial plans and intervention can be initiated by either party. On the other hand, \textsc{PGPlanner} is responsible for only acquiring expert knowledge wherever it is expected to be most useful. % in order to guide the plan search. 
\textsc{PGPlanner} builds on the HTN planning framework 
%Our work, builds on HTN planner as the underlying planning system that
since it allows the human to encode coarse knowledge and subsequently learn fine-grained knowledge via interaction.

\paragraph{HTN Planning:}
%\noindent {\bf HTN Planning Framework:} 
An HTN planner \cite{ghallab2004automated}, one of the well-known neo-classical planners, searches for valid plans in the space of tasks and sub-tasks by recursively decomposing the current task into sub-tasks based on pre-defined control knowledge, called \textit{methods}, and adding the sub-tasks into the current set. If a primitive task is solvable by an atomic action, it is removed from the set of tasks, the action is added to the plan, and the state is changed. Other \textit{non-primitive} tasks are then decomposed further. The resultant network of decompositions is a task network. Formally,
\begin{defn}
    A task network is directed acyclic graph $\mathbb{W}=(N,E)$. A directed edge $e=\langle \tau, \tau^{(sub)} \rangle$, where $\tau,\tau^{(sub)} \in N$ and $e \in E$, is always from a task $\tau$ to one of its sub-tasks $\tau^{(sub)}$ ( i.e. $\tau^{(sub)} \in subtasks(\tau)$).
\end{defn}
\begin{defn}
    A method is a tuple $m_\tau=(\tau, \mathcal{F}(a,s), \{\tau_j\}_{j=1}^k)$ where $\tau$ is the task for which $m_\tau$ is applicable, $\mathcal{F}(a,s)$ ensures $m_\tau$ is admissible (when $s$ satisfies $a$) in current state $s$ ($a$ is some admissibility criteria) and $\{\tau_j\}_{j=1}^k$ is the set of sub-tasks to which the task $\tau$ will be decomposed on application of $m_\tau$.
\end{defn}
Note that the above definition formalizes admissibility in a generalized fashion, independent of representational syntax. In HTN domain descriptions, however, admissibility of methods is represented as preconditions or a first order conjunctive formula that the current state $s$ must satisfy.

\begin{defn}
    A hierarchical task network problem is defined as  $P = (s_0, w_0, O, M)$ where $s_0$ is the initial state, $w_0$ is the initial task network, $O$ is the set of operators (atomic actions) and $M$ is the set of decomposition methods.
\end{defn}
As an intuitive example of how HTNs facilitate human experts in providing knowledge/feedback, consider building a house. The primary task is ``Build House'' which can be decomposed into subtasks: \textit{``Build House'' $\rightarrow$ [``Build Foundation'', ``Build Walls'', ``Build Roof'']$^1$}. Again the subtask ``Build Foundation'' can be decomposed further: \textit{``Build Foundation'' $\rightarrow$ [``Dig x ft'', ``Lay Reinforcement Bars'', ``Pour Concrete'']$^2$} (superscripts indicate levels of decomposition). Methods guide how such tasks need to be decomposed. Clearly, the varying levels of task abstraction allow for a human to provide feedback at different levels of generality. For instance, a human could say ``Base needs to be deeper than 10 ft'' at \textit{``Dig x ft''} level (level 2) or (s)he could also say ``Foundation must be $1/3^{rd}$ the height of the house'' at the higher \textit{``Build Foundation''} task level (level 1).

\paragraph{Preference Elicitation:} %Recent work in machine learning has focused on how to effectively use 
There has been a surge in the interest for using
human experts to improve decision-making~\cite{sohrabi2008planning,baier2009htn,Kunapuli13,judah2014imitation,odom2016active}.
While distinct methods differ in the form of knowledge that can be specified, preference elicitation has been explored inside automated planning~\cite{myers1996advisable,boutilier1997constraint,huang1999control,brafman2005planning,sohrabi2008planning}, in reinforcement learning (RL)~\cite{maclin1994incorporating,Maclin1996,torrey2005using,natarajan:2005} and inverse reinforcement learning (IRL)~\cite{Kunapuli13,odom2016active}. 
%Our work may be the closest in spirit to HTN-planning with preferences~\cite{baier2009htn}, but there are several fundamental differences which we clarify in the later sections. 

The most similar preferences to those used in our approach are IF-THEN statements where the IF defines the conditions (without negation) under which the preferences should apply and the THEN represents the preference. In RL or IRL, preferences could represent sets of preferred/non-preferred actions in a given set of states~\cite{torrey2005using,KunapuliEtAl10}. In HTN planning, preferences could correspond to preferred/non-preferred methods for decomposing certain tasks~\cite{baier2009htn}. Across these approaches, preferences have shown to be a good choice for specifying expert knowledge. 

However, many of these approaches require all of the preferences/advice upfront. This requires the domain expert who may not be a machine learning expert to provide the knowledge that would be useful for the learner. \textit{Alternatively, our approach solicits preferences during planning only as needed. This allows for effective interaction by reducing the number of uninformative queries}.

Recent work on active advice-seeking~\cite{odom2016active} introduces a framework to allow the learning algorithm to request preferences over interesting areas of the feature or state space. To the best of our knowledge, our approach is the \textit{first to actively solicit preferences in the planning setting}. 

Key contributions: Unlike active advice-seeking, planning does not have training examples from which to generalize. Instead, we select points to query during the planning process where the preferences may have the most impact. We show that even when learning without examples, an active learning framework can guide the learner towards effective and efficient communication with domain experts. 

%Recently, Odom et al.'s work on active advice seeking in Inverse Reinforcement Learning~\cite{odom2016active}, presented a framework to actively query for advice from human on uncertain states. Preference based HTN planning  \cite{baier2009htn} relate closely to the HTN based framework we adopt, but differ on how such preferences are acquired, encoded and fundamentally influence the search space. Our work is, sort of, an initial step towards a robust human-agent collaborative planner as conceptualized by  Allen et. al.\shortcite{allen2002human}. 

% --------------MAIN SECTION --------------------------------
%\vspace{-1em}
\section{Active Preference-Guided Planning}

Preference-guided planning employs preferences to guide the search through the space of possible plans, a list of primitive actions.
We make use of HTNs to search through these plans by recursively breaking a higher-level task into sub-tasks until every task can be solved using  primitive actions.
Our work differs from prior research on preference-based planning~\cite{sohrabi2008planning} in two distinct ways: 1) Our preferences are not used to define the best plan. Instead, they guide the decomposition of the network to efficiently find  high-quality plans; and 2) We aim to actively acquire preferences as needed during the search process as opposed to requiring the preferences upfront. Our preferences are defined as:

\begin{defn}
\label{defn:pref}
    A user-defined preference is defined as a tuple $\mathfrak{P} = (\wedge f_i, \tau_j, M_{\tau_j}^+,M_{\tau_j}^-$), where $\wedge f_i$ corresponds to conditions of the current state under which the preference should be applied\footnote{We use $\land f_i$ to denote that this could be a set of multiple conditions}, $\tau_j$ is the relevant task, $M_{\tau_j}^+$ is the set of methods which are in the  user's preferred set and $M_{\tau_j}^-$ is the set of methods which are in the user's non-preferred set.   
\end{defn}
A preference can be considered an IF-THEN rule where $\wedge f_i$ corresponds to the conditions of the current state under which the preference should apply to a particular task, $\tau_j$, and $M_{\tau_j}^+/M_{\tau_j}^-$ represents the method(s) preferred/non-preferred by the user for decomposition. It is important to note that a preference may be defined for (1) all instances of a particular task ($\wedge f_i=true$), (2) only a single instance of a task or, (3) any level of abstraction in-between. 

%How do we say the preference could represent an order over the methods or a plan to solve the subtask?

\begin{figure}[ht]
    \centering
    \includegraphics[width=.90\columnwidth]{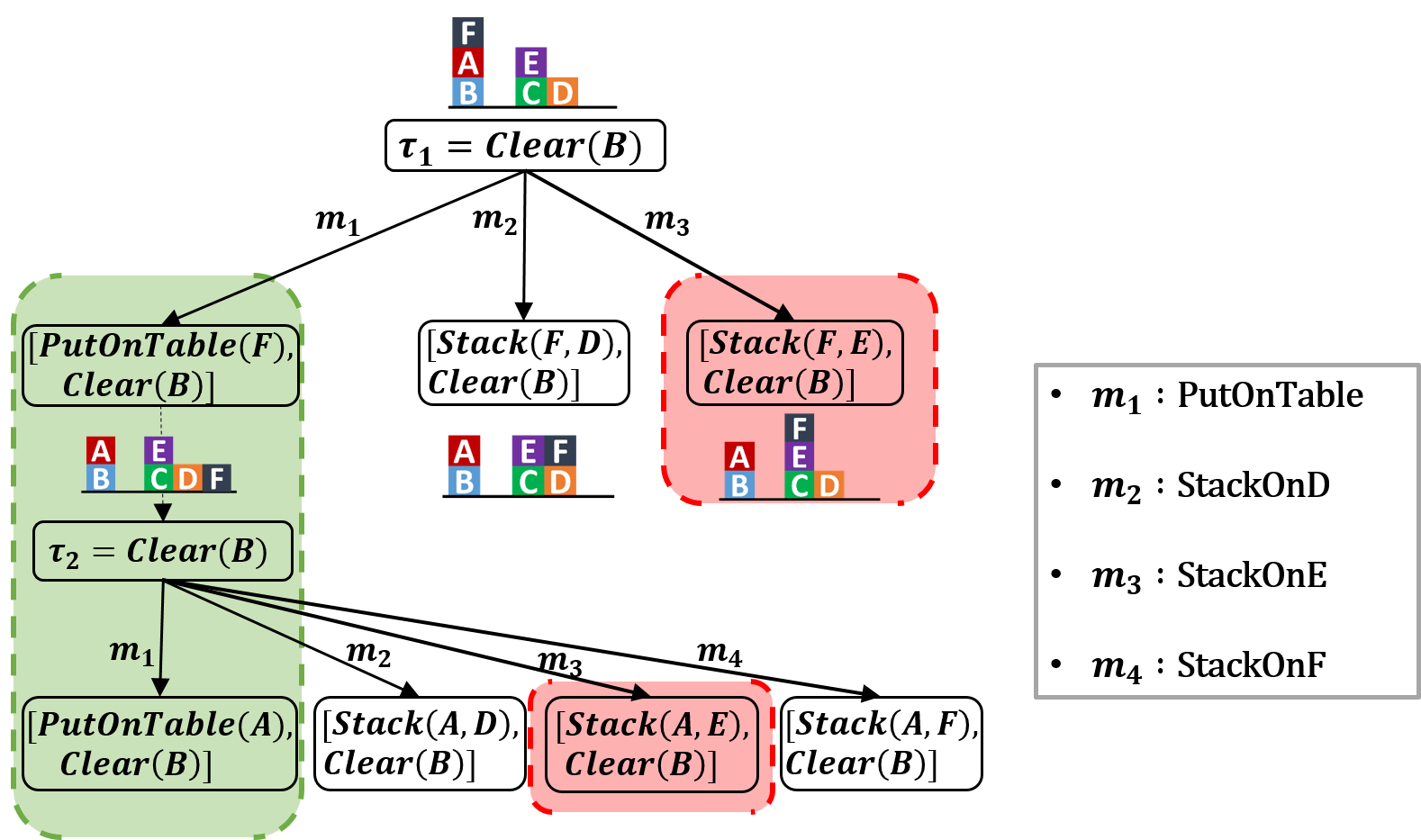}
    \caption{Preference guided search in a Blocks world problem. Rectangular nodes signify a task or a set of tasks to be solved. The admissible methods for decomposing the task $\tau_1$ are $m_1$, $m_2$ and $m_3$. Note how in the lower sub-tree we have an additional admissible method $m_4$. Block configuration diagrams signify the current state at every step. The green and red shaded areas denote preferred and non-preferred decompositions (best viewed in color).}
    \label{fig:prefplan}
\end{figure}

Our approach uses these preferences to guide the search through the space of possible decompositions in the HTN. Consider the network shown in Figure~\ref{fig:prefplan}. Each node in the network is labeled by the current state - the current configuration of the blocks - and task. For example, the root node represents the task $\tau_1$ of clearing block $B$ in the state where block $F$ is on $A$, $A$ is on $B$, etc. Note that we use predicate notation internally to represent the states. The edges in the network represent decompositions
%\footnote{In general, a task may decompose into multiple subtasks.} 
and are labeled with the method name. Method $m_1$ breaks task $\tau_1$ into the operator $PutOnTable(F)$ and, recursively, the task $clear(B)$.
%The edge width corresponds to the planners current distribution over the methods with thicker edges being more likely to be explored. 

\begin{eg}
\label{eg:expref}
    Figure~\ref{fig:prefplan} represents a preference $$\mathfrak{P} = (Space(Table), Clear(B),\{PutOnTable\}, \{StackonE\})$$ in Blocks World. 
    %and its impact on the search space of decompositions for an example problem in the Blocks World domain. 
    Shaded green areas represent preferred decomposition while shaded red areas represent non-preferred decompositions. %The preference is used to guide the search through the decompositions.   
    %({\bf Sriraam: Explain this better. What does it prune? Why is this advantageous? More importantly, why cannot this be specified up front?})
\end{eg}
The preference represents the intuition that it is beneficial to build towers of blocks when all the blocks are on the table as they can be positioned quickly. As specified, this single preference can apply at multiple points during the search. The effect of the preferences is to update the distribution that increases the probability of the methods $M^+$ and decreases the probability of the methods $M^-$. The subtree for $M^+$ is highlighted in green while the subtree for $M^-$ is shown in red. While a similar preference can be given upfront, our active approach evaluates whether this preference is necessary by estimating the quality of each method. Therefore, our approach increases the value of each preferences by reducing redundancy that may be present in upfront preferences.  

In the context of acquiring preferences from the expert, although we may  assume availability of the expert throughout the planning process, we aim to rely on him/her only when necessary. This setting is similar to stream-based active learning~\cite{Freund1997ML} where examples are shown online and the algorithm must decide whether to query for a label for the example or ignore it. Instead of acquiring a label, our preferences are more general allowing the expert to prefer/non-prefer methods for decomposition. 
A query is solicited over the current state $s_n$ and the current task $t_n$,
\begin{defn}
A query is defined over an HTN node $n$ as a tuple $q_n = (s_n, \tau_n)$.
\end{defn}
An expert's response to a query is a preference. As HTNs are hierarchical, the expert is not restricted to providing preferences only over the current state/task. It could also be defined over any subset of the state space that contains $s_n$. The expert selects the proper generality of the preference. When the space is factored, this involves removing features or introducing variables in description of $s_n$.

\begin{eg}
    In Figure~\ref{fig:prefplan}, $Clear(B)$ at the root node has $3$ different decomposition choices all of which may seem equally valuable to \textsc{PGPlanner}. The following query would be generated:
%    \begin{footnotesize}  
\begin{align*}
q = (<Space(Table), On(A,B) \wedge On(F,A), ... >,
 \tau=Clear(B))
\end{align*}
The expert may provide the preference specified in Example~\ref{eg:expref}. Note that in that case, the expert gives a general advice that applies to any state in which there is space on the table.
\end{eg}

%We need to include an example here
%%\begin{exmp}
%%    Consider a Blocks World domain where states are represented as conjunctions of ground atoms of the form $on(Block_1,Block_2)$, $ontable(Block_k)$ and $clear(Block_j)$ signifying some block $Block_1$ is not top of some $Block_2$, $Block_k$ is on the surface of the table and there is nothing on top of $Block_j$ respectively. The set of operators $O=$ \{$pickup(Block_i),$ $putdown(Block_i),$ $stack(Block_i,Block_j),$ $unstack(Block_i)$\}. Primitive tasks have the same structure and name as the operators. Non-primitive tasks can however be 
    
%%    In this scenario a user-defined preference $pref$, as defined in Definition~\ref{defn:pref},  can be 
%%    \begin{eqnarray}
%%    \nonumber pref & = & (on(Block_i,Block_j)\wedge clear(Block_k),\\
%%    \nonumber &  & \tau , M^+ = [unstack(Block_i),\\
%%    \nonumber &  & stack(Block_i,Block_k)])
%%    \end{eqnarray}
%%\end{exmp}

\begin{algorithm}[htbp]
\begin{algorithmic}[1]
\begin{small}
%\Procedure{PGPlanner}{$s_0,w_0,O,M$}
%\State $Frontier \gets$ all nodes in $w_0$ ordered by depth
%State $Plan = \emptyset, \mathfrak{P} = \emptyset, s=s_0$ \Comment{$\mathfrak{P}$ denotes preference set}
%\While{$Plan = \emptyset$}
%\State $n$ = \Call{Pop}{$Frontier$}
%If{$\tau_n$ is \textit{non-primitive}}
%\State $P(M_{\tau_n}), U(M_{\tau_n}) \gets$ \Call{EvalNode}{$n,\mathfrak{P}$}
%\State $\mathfrak{P} \gets \mathfrak{P}$ $\cup$ \Call{Query}{$U(M_{\tau_n}),n$}
%\If{$\mathfrak{P}$ has changed}
%\State $P(M_{\tau_n}), U(M_{\tau_n})$ $\gets$ \Call{EvalNode}{$n,\mathfrak{P}$}
%\EndIf
%\State $M^* \gets \underset{m}{argmax}$ $P(M_{\tau_n})$ \Comment{$m\in M_{\tau_n}$}
%\State $\{n^\prime\} \gets$ \Call{Decompose}{$\tau_n,M^*$}
%\State \Call{Push}{$Frontier,\{n^\prime\}$}
%\Else
%\State $apply(s_n,a_{\tau_n})$ \Comment{Add primitive to current plan}
%\EndIf
%\EndWhile\\
%\Return{$Plan$}
%\EndProcedure

\Procedure{PGPlanner}{$s_0,w_0,O,M$}
\State $Frontier \gets$ all nodes in $w_0$, $Plans = \emptyset, \mathfrak{P} = \emptyset$ 
\Statex \Comment{$\mathfrak{P}$ denotes preference set}
\While{$Plans == \emptyset$ and $Frontier \neq \emptyset$}
\State $currPlan \gets$ \Call{RecurSrch}{$\emptyset,Frontier$}
\If{$currPlan \neq \emptyset$ and $currPlan \neq NULL$}
    \State $Plans \gets Plans \cup currPlan$
\EndIf
\EndWhile
\State \Return{$Plans$}
\EndProcedure

\algrule
\Procedure{RecurSrch}{$currPlan,Frontier$}
\If{$Frontier \neq \emptyset$}
    \State $n$ = \Call{Pop}{$Frontier$}
    \If{$\tau_n$ is \textit{non-primitive}}
        \State $\pi(M_{\tau_n}), U(M_{\tau_n}) \gets$ \Call{EvalNode}{$n,\mathfrak{P}$}
        \If{\textbf{\textit{Not}} \Call{AcceptableUncertainty}{$U$,$n$}}
	    \State $\mathfrak{P} \gets \mathfrak{P}$ $\cup$ \Call{QueryExpert}{$m,s$} 
	    \Statex \Comment{Generate query and acquire preference}
        \State $\pi(M_{\tau_n}), U(M_{\tau_n})$ $\gets$ \Call{EvalNode}{$n,\mathfrak{P}$}
        \EndIf 
%        \State $\mathfrak{P} \gets \mathfrak{P}$ $\cup$ p %\Call{Query}{$U(M_{\tau_n}),n$}
%        \If{$\mathfrak{P}$ has changed}
%            \State $P(M_{\tau_n}), U(M_{\tau_n})$ $\gets$ \Call{EvalNode}{$n,\mathfrak{P}$}
%        \EndIf
        \State $M_{cur} = M_{\tau_n}$
        \While{$M_{cur} \neq \emptyset$}
            \State $M^* \gets \mbox{arg}\max_{m} \pi(M_{cur})$ \Comment{$m\in M_{curr}$}
            \State $\{n^\prime\} \gets$ \Call{Decompose}{$\tau_n,M^*$}
            \State $NewFrontier \gets$\Call{Push}{$Frontier,\{n^\prime\}$} 
            \Statex \Comment{Temporary frontier stack}
            \State $retVal \gets$ \Call{RecurSrch}{$currPlan$, $NewFrontier$} 
            \Statex \Comment{Recursive call}
            \If{$retVal \neq NULL$}
                \State \Return{retVal} \Comment{Backtracking}
%                \State \textbf{break}
            \EndIf
%            \If{$M_{\tau_n} = \emptyset$}
%                \State \Return{NULL} \Comment{Backtracking}
%            \EndIf
            \State $M_{cur} \gets M_{cur} - M^*$
        \EndWhile
%        \State \Return{NULL} \Comment{Backtracking}
    \Else
        \If{$Success\left(apply(currPlan,s_n,a_{\tau_n})\right)$} 
        \Statex \Comment{Applying primitive action to current plan}
            \State \Return{$currPlan$} %\Comment{Backtracking}
        \EndIf
    \EndIf
\EndIf
\State \Return{NULL} \Comment{Backtracking}
%\State \Return{$currPlan$} 
\EndProcedure

\algrule
%\Procedure{Query}{Uncertainty $U$, HTN node $n$}
%\If{\Call{AcceptableUncertainty}{$U$,$n$}}
%	\State $p =$ \Call{QueryExpert}{$m,s$}
%\EndIf \\
%\Return{$p$}
%\EndProcedure
%\algrule
\Procedure{EvalNode}{HTN node $n$, Preference $\mathfrak{P}$}
\State $\hat{C}=\emptyset$ \Comment{Set of scores $\forall m : m\in M_{\tau_n}$}
\For{\textit{\textbf{each}} $ m \in M_{\tau_n}$}
\State Node $r_m \gets$ \Call{Rollout}{$m,n,d$} 
\Statex \Comment{rollout depth $d$ is set to a constant}
\State $L_m \gets cost(plan_{r_m})$, $D_m \gets \delta(s_{r_m}, goal)$
%\State $A \gets \left|\{p|p\in\mathfrak{P},m \in M^+_p\}\right| - \left|\{p|p\in\mathfrak{P},m \in M^-_p\}\right|$
\State $\mathfrak{P}(s_n) \gets \forall_{p \in \mathfrak{P}}$ $(\wedge f^j_i \wedge s_n = true) \wedge (\tau_p = \tau_n)$ 
\State $A_m \gets { N^+_m(\mathfrak{P}(s_n)) - N^-_m(\mathfrak{P}(s_n) })$ 
\State $\hat{C} \gets \hat{C} \cup \langle m, ( (D)^{-1} + (L)^{-1}  +  A) \rangle$ 
\Statex \Comment{Maximize adherence \& minimize cost}
\EndFor
\State Compute $\pi(M_{\tau_n})$  %\Comment{using $S(M_{\tau_n})$.}
\State Compute $U(M_{\tau_n}) \gets {\sum}_{m\in M_{\tau_n}} p(m) . \log (1/p(m))$ 
\Statex \Comment{Entropy from $\pi$}
\State \Return{$\pi(M_{\tau_n}),U(M_{\tau_n})$}
\EndProcedure
\end{small}
\end{algorithmic}
\caption{Preference-Guided Planning}
\label{algo:pgp}
\end{algorithm}

\subsection{Problem Overview}
%\begin{align*}
%    \textbf{Input:} HTN $\mathcal{H}$, Initial Preferences (may be empty), Budget on queries $k$, Goal Task, Expert \\
%    \textbf{Output:} Plan achieving the goal task
%\end{align*}
%In our work, we assume the existence of a metric which measures the optimality of a particular plan, i.e. the shortest plan. However, the planner is not aware of this metric during planning. 
\textsc{PGPlanner} finds a plan given an HTN defining the initial state/goal task(s), and access to an expert. The goal of \textsc{PGPlanner} is to find the policy $\pi$, a distribution over the methods for each HTN node $n$, such that the best plan is reached:
\begin{equation}    
\mbox{arg}\min_{\pi} \; \left( J(\pi)=T \mathbb{E}_{n\sim d_\pi} C_\pi(n) \right)
\end{equation}
$J(\pi)$ is the total expected cost of finding a plan. If $\pi$ is used to select decompositions, $d_\pi$ represents the distribution of HTN nodes reached. $T$ is the depth of the decomposition. $C_\pi(n)$ is the expected cost at node $n$. $C_\pi(n) = \mathbb{E}_{m\sim \pi_n} C(n,m)$ where $C(n,m)$ is the immediate cost of selecting $m$ at node $n$. If the planner aims to find the shortest plan, then the immediate cost $C$ will be the number of actions added to the current plan.

\textsc{PGPlanner}, Algorithm~\ref{algo:pgp}, recursively searches through the space of possible HTN decompositions to reach a valid plan. Each node $n$ in the HTN with task $\tau_n$ could potentially decompose in several ways according to the available methods ($M_{\tau_n}$). The cost of selecting a method $m \in M_{\tau_n}$ ($C_{\pi}(n)$ for $\pi(n)=m$) is estimated
by rolling out the current plan and then approximating the distance to the goal.  The methods are also scored according to the current set of preferences. The overall cost estimate ($\hat{C}(m)$) is a combination of this preference score and the estimated cost function. Finally, this is converted into a probability distribution ($\pi$) over the methods $M_{\tau_n}$. If this distribution has a high-level of uncertainty (entropy in our case), the expert is queried about the current set of possible methods.

%The \textsc{PGPlanner} maintains the set of all HTN nodes that could be explored. As \textsc{PGPlanner} uses a best-first search, the $Frontier$ stores the nodes ordered by their deptreh. All nodes of the same depth are then ordered by their estimated scores (described later). The planner will return the first plan found.

\subsection{The algorithm}
The \textsc{PGPlanner} maintains a {$Frontier$}, the set of all HTN nodes that have to be explored. It is initialized with 1 or more nodes containing the goal task(s). \textsc{PGPlanner} proceeds by recursively decomposing the task $\tau_n$ of the node $n$ at the head of the frontier and inserting new nodes for the sub-tasks of $\tau_n$.
The methods of non-primitive tasks are recursively selected based on $\hat{C}(m)$ (lines \textbf{21-30}). Primitive tasks are solved by adding the operator to the current plan (line \textbf{32}). This apply step updates the plan for all ancestors of the current node.

When evaluating a node $n$ of the HTN (\textsc{EvalNode}), the methods $m \in M_{\tau_n}$ represent the set of possible choices. We estimate the cost $\hat{C}(m)$ for each method by rolling out for $d$ steps. $L_m$ represents the estimated cost of the roll-out on method $m$. In our case it corresponds to the plan length. $D_m$ approximates cost to reach the goal state from the state after the roll-out is completed (line \textbf{43}). This distance, denoted as $\delta$, is the number of unsatisfied goal atoms in the current state. Along with the estimated cost, methods are also evaluated with respect to the set of preferences by the adherence score ($A_m$). 
This score (line \textbf{45}) is  determined by the number of preferences applying to the current node $n$ ($\mathfrak{P}(s_n)$). The number of preferences which prefer method $m$ is represented by $N^+_m$ while $N^-_m$ represents the number of preferences which non-prefer it. Notice that this formulation could allow conflicting preferences on a single method and task. The final score ($\hat{C}(m)$) is a combination of the estimated cost ($L_m, D_m$) and the adherence score (line \textbf{46}). We convert this score into a distribution ($\pi(n)$) over the methods using a Boltzmann distribution, $p(m) = e^{\hat{C}(m)}/\sum_{x\in{M_{\tau_n}}} e^{\hat{C}(x)}$ where $m\in M_{\tau_n}$ and $p(m) = \pi(n,m)$. This distribution is used in two ways: to select the exploration order of the methods and to decide whether a query is necessary.

The query decision is based on the uncertainty over the set of possible methods (lines \textbf{16-19}). %While we focus on the uncertainty, other criteria like the current depth of the node or the current task could also be used. For example, queries may be more useful near the root of the search tree as they influence the options as the problem is further decomposed. 
Inspired by the success of Active Learning\cite{settles10}, we use the uncertainty measure to query the expert. However, one could replace this with any function that needs to be optimized - cost to goal, depth from the start state or a domain specific utility function etc. to name a few. \textsc{PGPlanner} uses entropy computed from $\pi(n)$ as the measure of uncertainty. Our framework uses a threshold on the entropy as the \textit{\textbf{Not}} \textsc{AcceptableUncertainty} to initiate a query. The expert can provide decomposition preferences over all tasks of a given type, specify preferences over a single instance of a task, or even provide a plan to solve a particular subtask. This provides an expressive framework for the expert to interact with the planner. The interaction is driven by the planner, allowing it to only ask as needed.

Overall, \textsc{PGPlanner} interacts with the expert to guide the search through the space of possible plans. This allows our algorithm to learn potentially better plans in a more efficient manner. We now briefly analyze some of the properties of \textsc{PGPlanner}.  

%\paragraph{Properties of the Proposed Approach:} 
\paragraph{Difference in obtaining preference at various steps:} First, we aim to quantify getting preference in earlier steps when compared to getting preference at later steps. Let us denote the probability of choosing a method according to the optimal plan as $p^o(m)$ given by the Boltzmann distribution. Recall that the cost of selecting method $m$ at HTN node $n$ as $C(n,m)$ and the cost of a policy $\pi$ is $C_\pi(n) = \mathbb{E}_{m\sim \pi_n} C(n,m)$. Now if we use a boolean error function that is set when the method at node $n$ is not chosen according to the policy ($e(n,m)=I(m\neq \pi^*(n))$, then the error of the policy is $e_\pi(n) = \mathbb{E}_{m\sim \pi_n}e(n,m)$. 

Our goal is to minimize the total expected cost of a policy $\pi$, $J(\pi)=T\mathbb{E}_{n\sim d_\pi} C_\pi(n)$. Ross and Bagnell~\cite{ross2010efficient} have shown for any policy $\pi$, $J(\pi)\leq J(\pi^*) + kT\bar{\epsilon}$, where $\pi^*$ is the optimal policy, $T$ is the task horizon, $k$ is the number of steps to the goal, and $\bar{\epsilon} = \frac{1}{T}\sum_i \epsilon_i$, where $\epsilon_i = \mathbb{E}_{n\sim d_{\pi^*}} e_\pi(n_i)$ is the expected error at node $i$.

A natural question is, does it benefit to ask the query early or should the planning algorithm wait to query the expert. This can be analyzed using the regret framework. 
Let $\pi_i$ and $\pi_j$ denote the policy $\pi$ when asking the query at steps $i$ and $j$ respectively ($j>i$).
%Let policy $\pi_i$ denote asking the query at step $i$ and $\pi_j$ at step $j$, where $j>i$. 
Now, rearranging terms, we can show that
%\begin{equation}
$J_{\pii} - J_{\pij} = \Delta(j-i)$,
%\end{equation}
where $\Delta$ denotes the expected change in error, if assumed to be the same in both the steps $i$ and $j$. 
%of choosing step $i$ vs $j$. 
Thus, the difference between the two choices to solicit preference is linear in the time difference between the two steps, and linear in the change in error in the two steps. Here it is clear that soliciting advice early can reduce the expected total cost.
%error difference between the two steps. 

%\vspace{-1em}

\paragraph{Benefit of Preference over rollout:} We now consider briefly analyzing the value of PGPlanner vs a simple rollout based planning. Let us denote the distribution over methods of optimal policy, PGPlanner, and rollout as $\pi^o$, $\pi^A$ and $\pi^R$ respectively, where $p^i(m) = \pi^i(n,m)$ where $m \in M_{\tau_n}$ is the posterior of choosing a particular method according to the policy. Suppose we compute the KL divergence between the probability distribution of methods of PGPlanner and rollout with the optimal distribution, 
\begin{align}
\nonumber   D_R & = D_{KL}(\pi^o||\pi^R)
    & = \sum_{m \in M(T)} p^o(m) . \ln \left(\frac{p^o(m)}{p^R(m)}\right)
\end{align}
\begin{align}
\nonumber    D_A & = D_{KL}(\pi^o||\pi^A)
    & = \sum_{m \in M(T)} p^o(m) . \ln \left(\frac{p^o(m)}{p^A(m)}\right)
\end{align}
This gives us the difference between $D_R$ and $D_A$. $D_R - D_A  =  \sum_{m\in M(T)} p^o(m) . log\frac{p^A(m)}{p^R(m)}$, which is simply a weighted sum of the log odds of the probability of choosing a method. It is possible to find the best $\pi^A$ that maximizes this difference by setting $\sum_m p^A(m)=1$ as a constraint, but this requires having access to the optimal distribution. Since that is unknown in many cases, one can simply observe that when the preferences drive the distribution over methods towards the optimal one, i.e., choose a method that is close to the optimal, the difference is $\geq 0$ indicating that the preference is more useful than the simple rollout. We next show empirically, this is indeed the case in many planning problems.

\begin{figure}[ht]
    \centering
    \includegraphics[width=0.9\columnwidth]{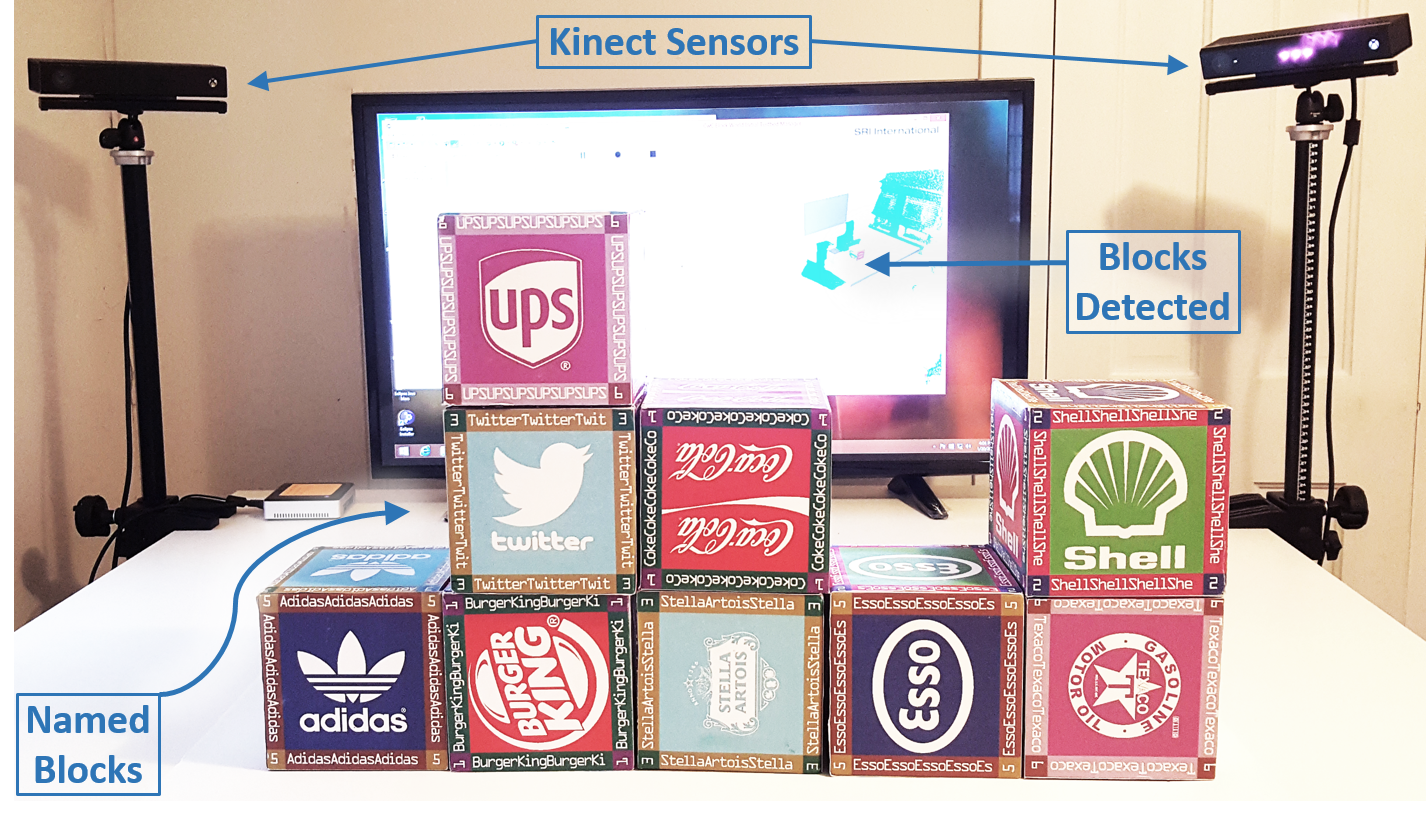}
    \caption{Blocks World Apparatus}
    \label{fig:apparatus}
\end{figure}

\begin{table}[ht]
    \centering
    {
    \begin{tabular}{|l|c|c|c|}
    \hline
    {{\small \textbf{Domains}}} & {\small $\mathbf{\#[Relations]}$} & {\small $\mathbf{\#[Objects]}$} & {\small $\mathbf{\#[Problems]}$}\\
    \hline
    Freecell & 5     & 52   & 20 \\
    %\hline
    Rovers & 27    & 50     & 20\\
    %\hline
    Trucks & 10    & 32     & 20 \\
    %\hline
    Depots & 6     & 45     & 20\\
    %\hline
    Satellite & 8     & 69  & 20\\
    %\hline
    TidyBot & 24    & 100   & 10\\
    
    Towers of Hanoi & 10 & 9  & 20\\
    
    Barman & 8 & 50 & 10\\
    
    Mystery & 12 & 35 & 10\\
    
    Assembly & 10 & 15 & 10\\
    
    Rockets & 6 & 15 & 10 \\
    \hline
    \hline
    Blocks World & 3     & 40   & 20\\
    \hline
    \end{tabular}}
    \caption{Experimental domains and their properties}
    \label{tab:domaindesc}
\end{table}
% --------EXPERIMENTS SECTION--------------------------------
\section{Experiments}
%--------------- Experimental figures ---------------------
\begin{figure*}[t]
    \centering
    %\subfigure%[Efficiency of plan generation]
    %{
        \includegraphics[width=\textwidth]{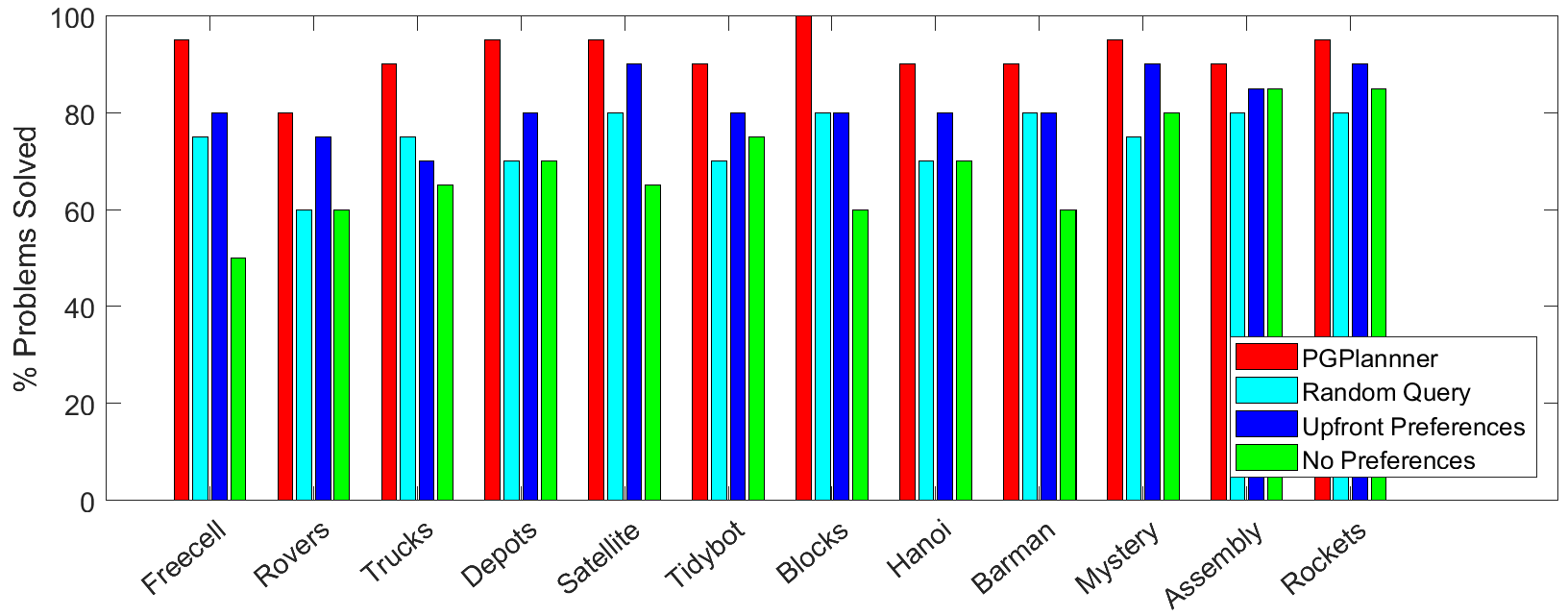}
        \caption{Efficiency comparison of all approaches across 12 domains. Percent problems solved in 10 minutes, higher is better. (best viewed in color)}
         
        \label{fig:percentproblems}
    %}\\
    %\subfigure%[Quality of plans]
    %{
    %\includegraphics[width=\textwidth]{PlanLength.png}
   
    %\label{fig:optimality}
    %}
    %\caption{Performance of all approaches across 12 domains. TOP, shows the percentage of problems solved in 10 minutes, higher implies better performance. BOTTOM, compares the ratio of average plan lengths for every approach to the longest average plan length, lower implies better (best viewed in color)}
    %\label{}
\end{figure*}

\begin{figure*}[t]
\centering
    \includegraphics[width=\textwidth]{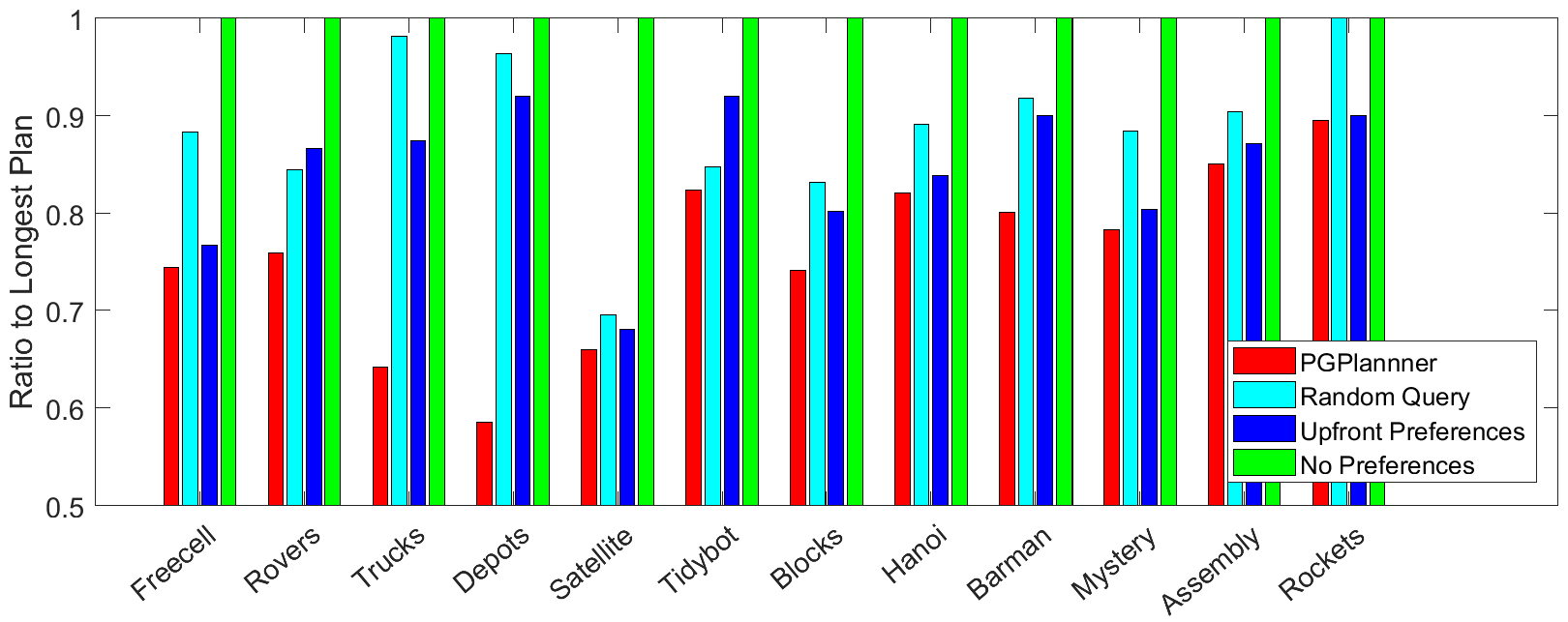}
   \caption{Performance comparison of all approaches across 12 domains. Compares the ratio of average plan lengths for every approach to the longest average plan length, lower implies better (best viewed in color)}
    \label{fig:optimality}
   % \label{}
\end{figure*}

\begin{figure*}[tb]
    \centering
    \subfigure[Freecell]{
        \includegraphics[width=.3\textwidth]{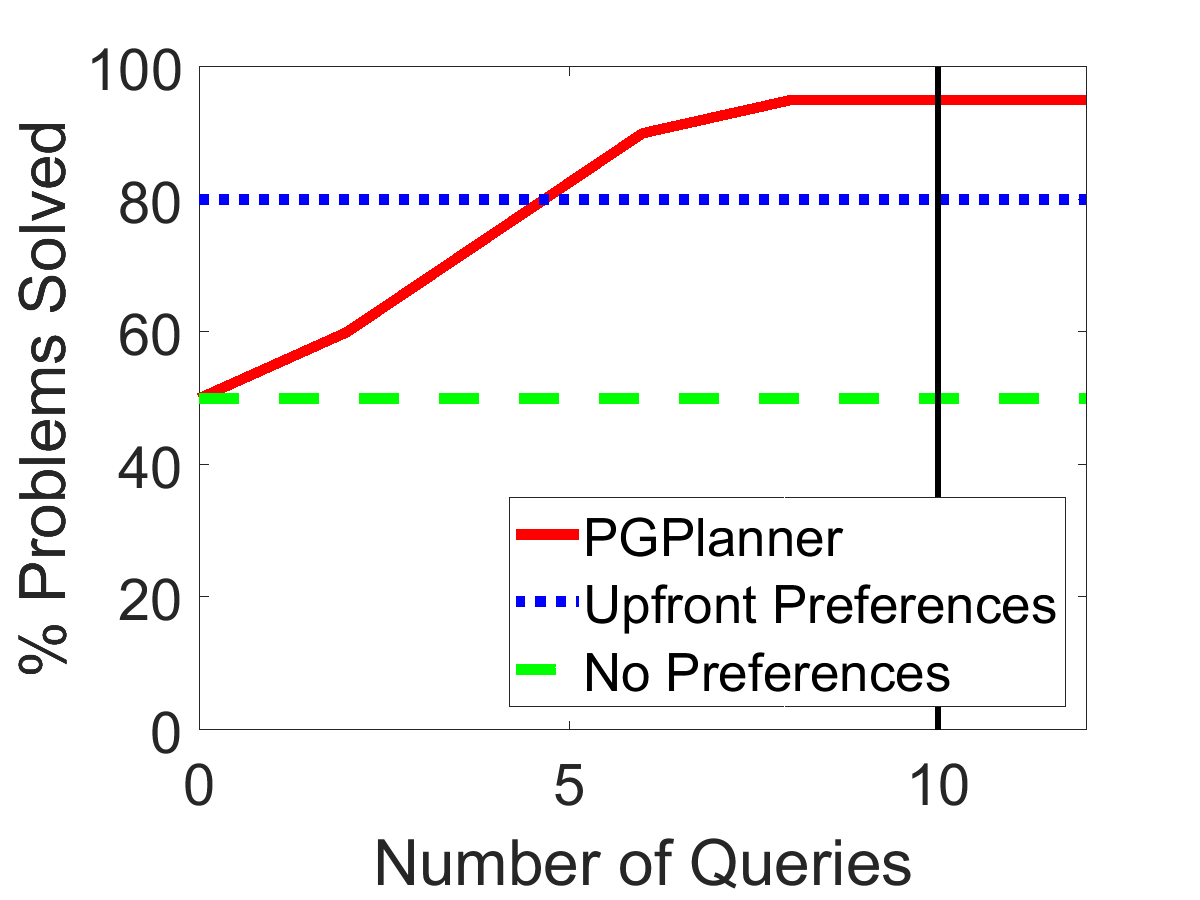}
        \label{fig:LCfreecell}
    }
    \subfigure[Blocks World]{
        \includegraphics[width=.3\textwidth]{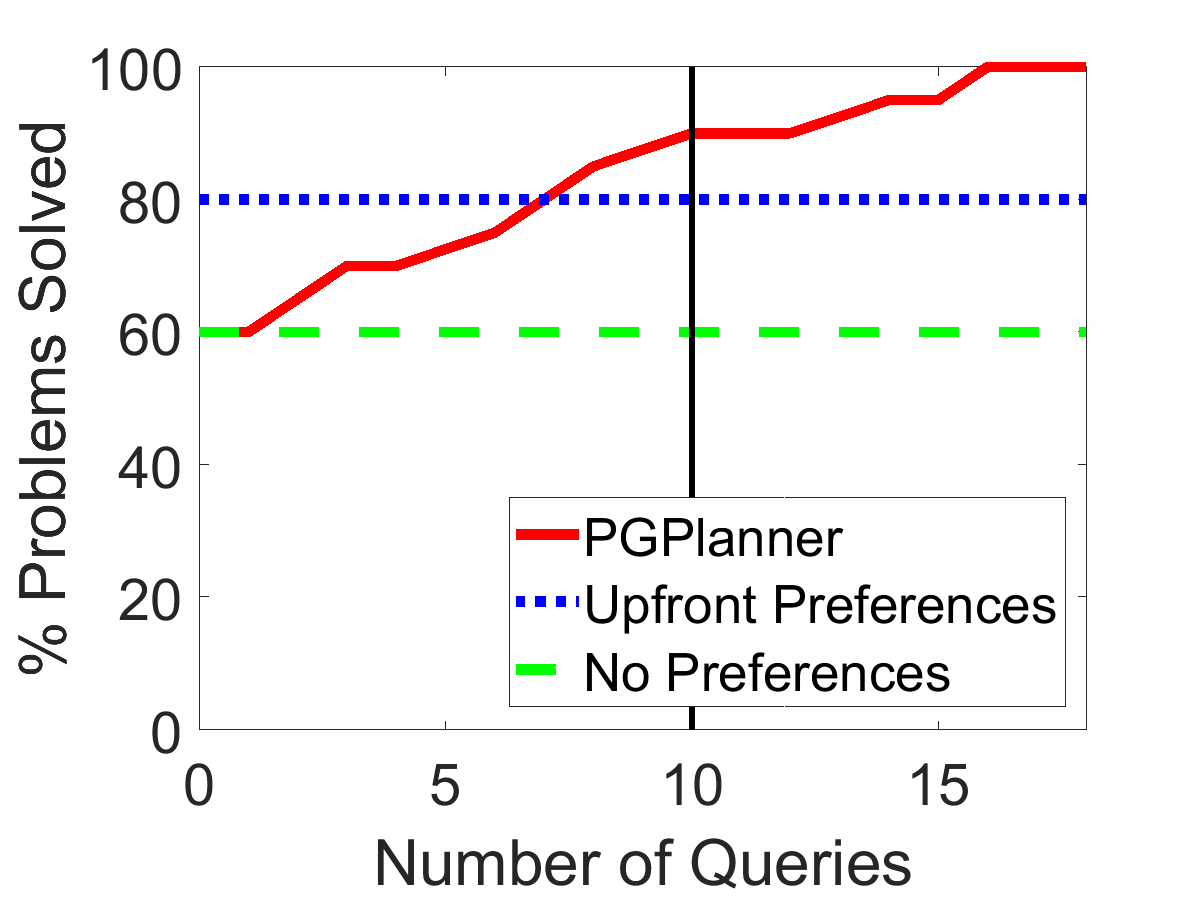}
        \label{fig:LCblocks}
    }
    \subfigure[Barman]{
        \includegraphics[width=.3\textwidth]{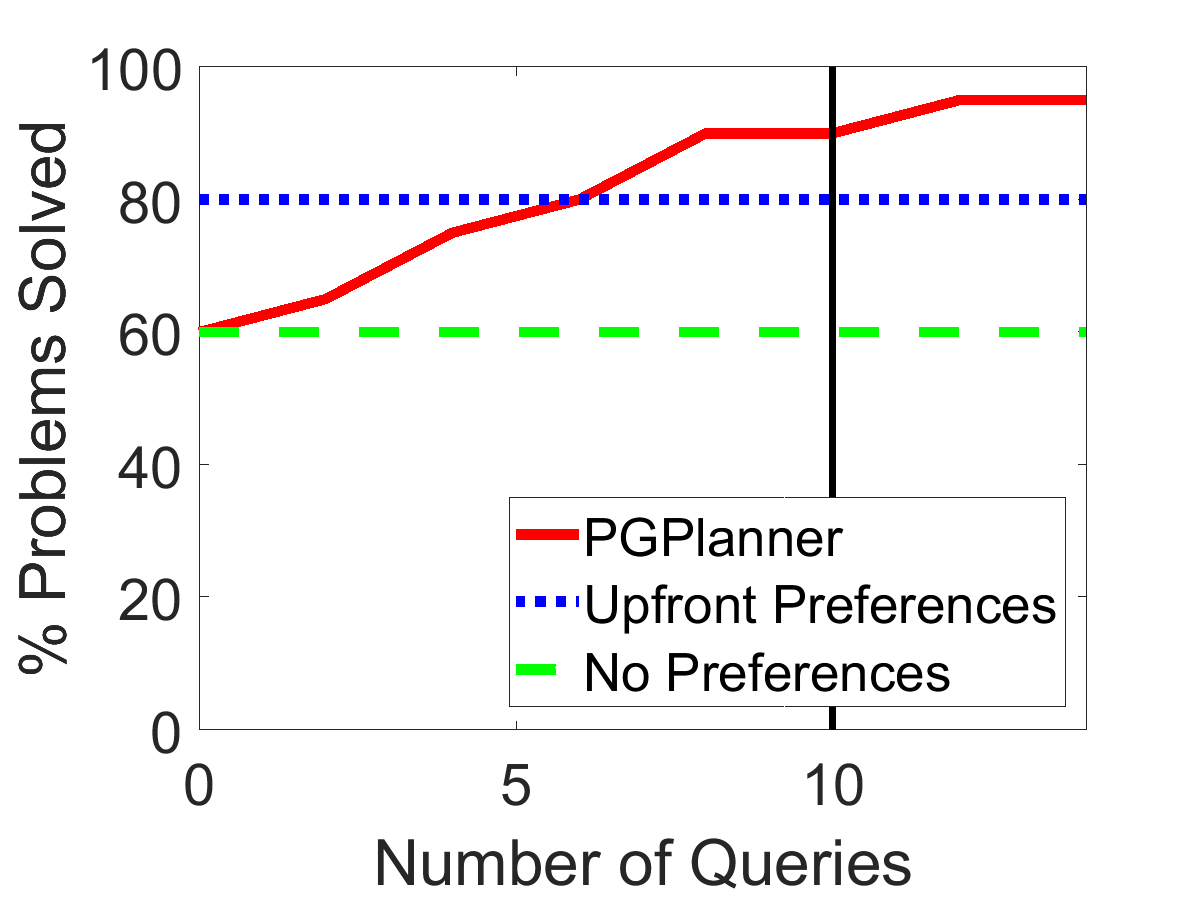}
        \label{fig:LCbarman}
    }
    \caption{Learning Curves in 3 construction domains (performance, \% of problems solved, vs. the number of queries.). The vertical lines denote the points where the number of preferences given upfront equals the number of queries by \textsc{PGPlanner} (best viewed in color).}
    \label{fig:LC}
\end{figure*}

\begin{figure*}[t]
    \centering
    \subfigure[Rovers]{
        \includegraphics[width=.3\textwidth]{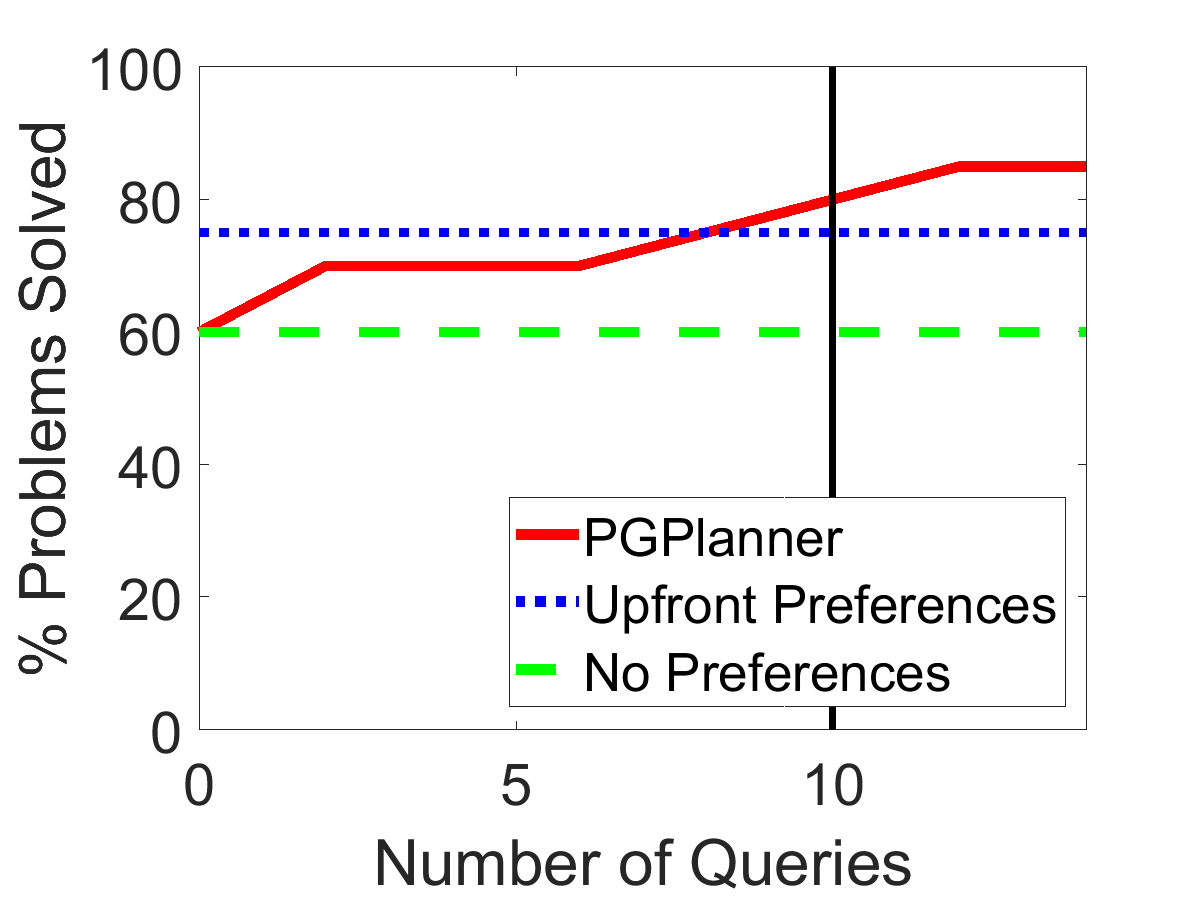}
        \label{fig:LCrov}
    }
    \subfigure[Depots]{
        \includegraphics[width=.3\textwidth]{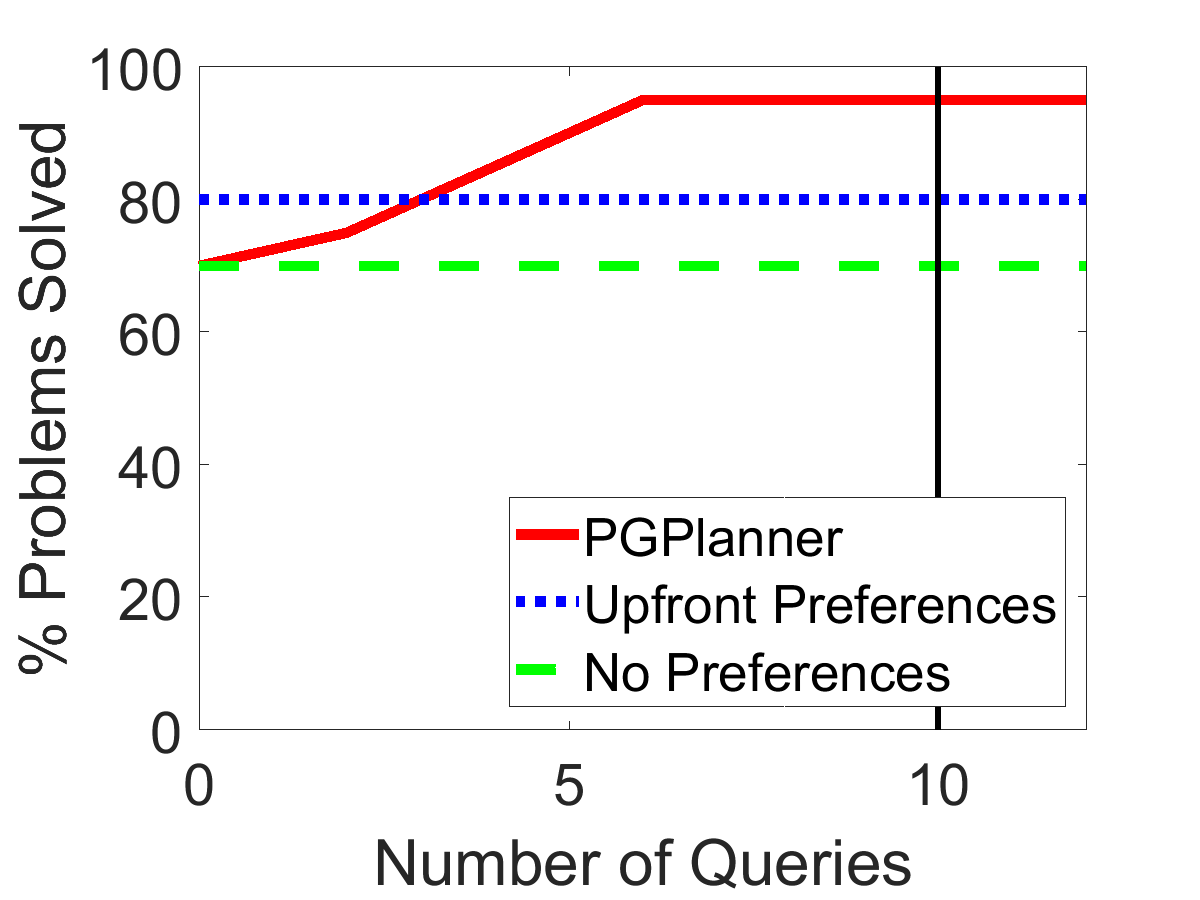}
        \label{fig:LCdepots}
    }
    \subfigure[Mystery]{
        \includegraphics[width=.3\textwidth]{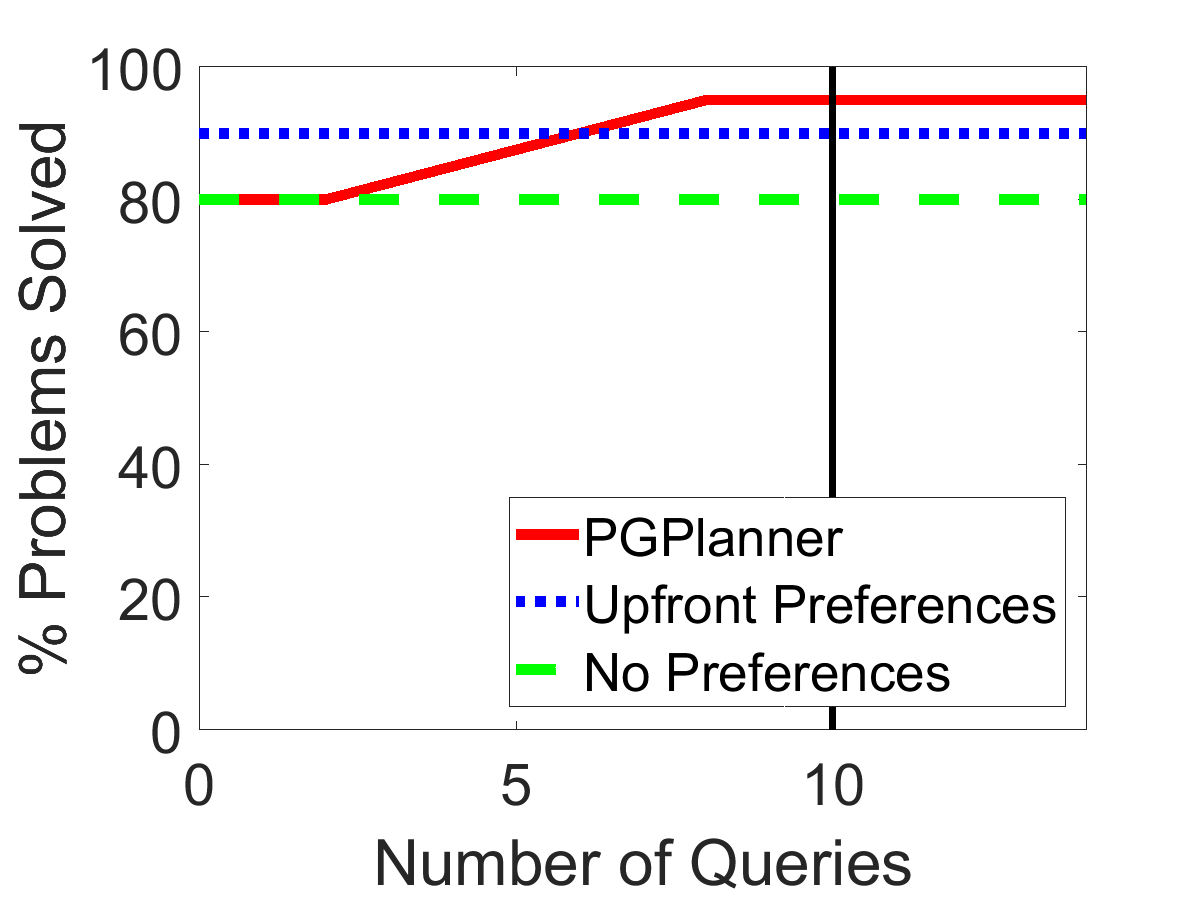}
        \label{fig:LCmys}
    }
    \caption{Learning Curves in 3 domains which focus on route finding and multi-location task solving (performance, \% of problems solved, vs. the number of queries.) (best viewed in color).}
    \label{fig:LC2}
\end{figure*}
%-----------------------------------------

%As discussed earlier our approach \textit{UnAct} presently builds on top of a HTN based planning framework. 
Our \textsc{PGPlanner} is built on top of the SHOP2 \cite{nau:shop2} architecture (called JSHOP), an HTN planner. We have extended the base planner to: (1) perform a roll-out to evaluate all admissible methods for decomposing the task,  (2) elicit human feedback/preferences based on our evaluation, and (3) utilize the preferences to guide the search. % However, original implementation of Sohrabi et. al.'s work \shortcite{baier2009htn} on preference based HTN is in \textit{lisp}, since the authors built on the original lisp implementation of SHOP2. We experimented with their original code (using the same problem sets) for baselines.
%\subsection{Experiments}

\paragraph{Empirical Research Questions.} Our experiments aim to answer the following questions, 
%\begin{description}
 %   \item
 [\textbf{Q1:}] Does \textsc{PGPlanner} generate plans efficiently? 
  %  \item
  [\textbf{Q2:}] How effective are the generated plans?
   % \item
   [\textbf{Q3:}] Does active preference elicitation improve the interaction with the expert?
%\end{description}

\noindent {\bf Baselines.} We compared \textsc{PGPlanner} against several alternate approaches for preference elicitation including (1) \textit{Upfront Preferences} - where all the preferences are specified before planning, similar to~\cite{baier2009htn}, (2) \textit{Random Query} - selecting whether to query randomly (for each step), and (3) \textit{No Preferences} - planning without preferences. In all of the experiments, we perform the role of the experts in providing preferences. Performing user studies is an interesting future direction.

We evaluate \textsc{PGPlanner} on several standard planning domains as well as a novel Blocks World domain, listed in Table~\ref{tab:domaindesc} (Number of relations, maximum number of objects and number of problems considered in each domain). There is no straightforward way to succinctly describe the complexity of the domains. However, the maximum number of objects and relations in each of them should provide a fair idea of its complexity. Experiments for the Blocks World were performed using a surrogate real-world environment, an apparatus (Figure~\ref{fig:apparatus}) that can detect block configurations of actual named blocks via sensors. State encoding is then generated by processing the sensor data. 

For all experiments, we set \textsc{AcceptableUncertainty} to \\$entropy$ $\leq 0.5$. We also performed line search on the value space, however, a threshold of $0.5$ worked well throughout and we report results with that. Preferences were provided by the person designing and conducting the experiments. We avoid experimental bias in the quality of preferences given across all the preference-based approaches. We verify this experimentally by storing all the actively elicited preference statements in a log file and using those as the set of input preferences in the Upfront approach and observing how they affect the decision making process in both cases (see Discussion).
%The second, third and the fourth column in the table signify the number of predicates (relations), the maximum number of objects considered and the number of problems, respectively, for each domain. 
% Can we have a short table to explain each domain and give key facts, number of problems, task description, size of domain...
%\begin{table}[t]
%  \centering
%  \caption{Experimental domains and their properties}
%  \scalebox{0.8}{
%    \begin{tabular}{|l|c|c|c|}
%    \hline
%    {\textbf{Domains}} & {$\mathbf{\#[relations]}$} & {$\mathbf{\max \left(\#[objects]\right)}$} & {$\mathbf{\#[problems]}$}\\
%    \hline
%    Freecell & 5     & 52   & 20 \\
%    %\hline
%    Rovers & 27    & 50     & 20\\
%    %\hline
%    Trucks & 10    & 32     & 20 \\
%    %\hline
%    Depots & 6     & 45     & 20\\
%    %\hline
%    Satellite & 8     & 69  & 20\\
%    %\hline
%    TidyBot & 24    & 100   & 10\\
%    \hline
%    \hline
%    Blocks World & 3     & 40   & 20\\
%    \hline
%    \end{tabular}%
%    }
%  \label{tab:domaindesc}%
%\end{table}%

%[h]
%    \centering
%    \subfigure[Blocks World Apparatus]{
%        \includegraphics[width=.55\columnwidth]{apparatus3.png}
%        \label{fig:apparatus}
 %   }
   % \subfigure[Our interface showing a Freecell problem]{
    %    \includegraphics[width=.4\columnwidth]{interface.png}
    %    \label{fig:interface}
    %}
%\end{figure}

%\nolinenumbers

%\linenumbers

We have developed an interface that facilitates the interaction between the expert and the planner. The interface has three main components: the state module, the partial plan module and the interaction module to visually render the current state, to expose the presently selected set of primitives and to provide a console for the human-agent interaction respectively.
%The state module shows the current state for the node in the HTN where the planner is currently exploring. Rather than a text description, it renders the state visually. The partial plan module gives the human expert some context in the current primitives already selected. 
The interaction module allows the planner to query the user and the user to respond to the query with a preference.
%\begin{figure*}
%    \centering
%    \includegraphics{ProblemsSolved.png}
%    \caption{Percentage of problems solved in a time constraint of 10 minutes in various domains}
%    \label{fig:percentproblems}
%\end{figure*}

%\begin{figure*}
%    \centering
%    \includegraphics{PlanLength.png}
%    \caption{Number of problems in which one approach generated better plans (lower cost) than the other. 3 distinct comparisons - \textit{PGPlanner vs No Preference, PGPlanner vs Random Querying and PGPlanner vs Upfront Preference}}
%    \label{fig:optimality}
%\end{figure*}

\subsection{Experimental Results}
In each domain, the planners were executed for 10 minutes. This allows for validating the ability of the expert to guide the algorithm to efficient solutions (as well as accommodating the limited time and attention of the expert). Figure~\ref{fig:percentproblems} shows the percentage of problems where a plan was found. We evaluate the quality of the learned plans separately. 

In most domains, every planner using preferences is able to outperform standard planning (no preferences). This indicates that preferences have a positive impact during planning. Planning with upfront preference outperforms randomly querying for preferences in 9 out of 12 domains. A disadvantage of random querying is that it may not query at points in planning where the preferences could have the most impact. Specifying preference upfront can take advantage of preferences at these crucial decisions at the cost of placing additional responsibility on the expert to give useful preferences. 
Across all domains, \textsc{PGPlanner} outperforms all of the baselines. This answers {\bf Q1} affirmatively in that actively eliciting preferences guides the planner to solutions more efficiently

%Actively eliciting preferences guides the planner to solutions more efficiently (\textbf{Q1}). 

Next, we investigate the generated plans. In every domain, we only compare problems where all planning methods are able to generate a plan (in the given time constraint). Figure~\ref{fig:optimality} illustrates the ratio of the average plan length of each planning method compared to the average of plans generated without preferences\footnote{The value for No Preference case is always $1$, since the ratio is taken with itself.}. Planning with preferences generates shorter plans in all the domains. Since, as evident, no preference always results in higher average plan lengths. The ratios for all methods with preferences are less than 1. However, \textsc{PGPlanner} has the lowest ratio across all domains. Thus, \textsc{PGPlanner} is able to produce shorter plans on an average than all of the baselines, thus answering \textbf{Q2} affirmatively as well.    
%How many of these are significant?

Finally, we investigate how our method performs relative to the number of queries solicited.  Figure~\ref{fig:LC} and ~\ref{fig:LC2} show learning curves, in 3 construction oriented domains (Freecell, Blocks World and Barman) and 3 route-finding and multi-location task oriented domains (Rovers, Depots \& Mystery), respectively. Note that learning with no preferences and learning with upfront preferences are constant as the number of preferences never changes. In each domain, the x-axis represents the number of preferences given by the expert. \textit{The vertical line denotes the point in the curve where the number of preferences given upfront equals the number of queries}. Our method outperforms learning with upfront preferences using the same number of preferences in all domains. This suggests that actively eliciting preferences succeeds in generating queries at important stages during the search process, improving the interaction with the expert (affirmatively answering \textbf{Q3}).

\subsection{Discussion}
\begin{figure}[h]
    \centering
    \includegraphics[width=0.9\columnwidth]{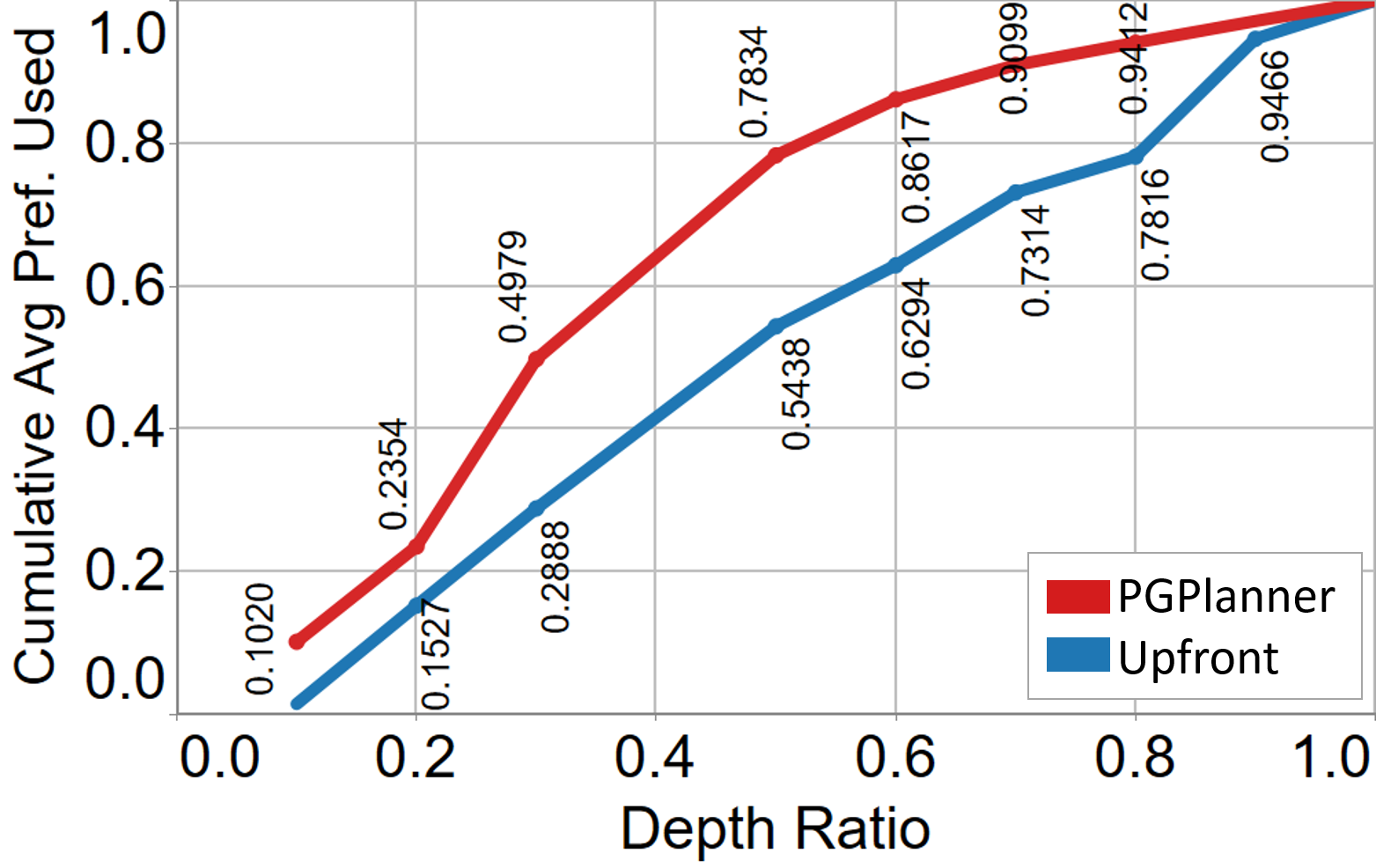}
    \caption{Preference used (cumulative) against relative depth (best viewed in color)}
    \label{fig:depth}
\end{figure}

\textsc{PGPlanner} is effective as the framework elicits preferences when and where they are necessary as well as relevant. Hence, the expert can provide informative preferences to steer the search process towards more useful parts of the search space.  Alternatively, upfront preferences may not always be relevant, because the preferences might be about unreachable regions, or may not alter the decision that the planner would have taken. We verify this by measuring the number of times the preferences were used during the search and how many of those alter the decision of the planner. We observe that on an average across all domains, upfront preferences are used at least 20\% fewer times (corresponding to 2.04 fewer uses) than preferences elicited by \textsc{PGPlanner}. 
%New Addition ------------------------
%Furthermore, actively acquired preferences influence the decisions in 90\% of the cases where the preference applies, compared to 78\% for upfront preferences. 
We evaluated how \textsc{PGPlanner} queries for (uses) preference across different stages/depths of plan search. Figure~\ref{fig:depth} shows how total average preferences used varies with relative depth. Since planning depth is different for every problem in every domain, 'depth ratio' denotes a standardized scale constructed by considering equidistant fractions of the total planing depth. Similarly, as total number of preferences used varies across domains, they were normalized and averaged over all domains, and their cumulative values were plotted. We observe how \textsc{PGPlanner} acquires 80\% of the preferences by 60\% of the planning depth as compared to the upfront case which uses preferences uniformly till completion. This empirical result corroborates the earlier theoretical contribution that showed the relationship between impact of the preference and the relative depth. 
%----------------------------------------------------------------

\begin{table}[ht]
    \centering
  {
    \begin{tabular}{|l|c|c|}
    \hline
    {{\textbf{Domains}}} & {\textbf{\textsc{ PGPlanner }}} & {\textbf{Upfront Preference}} \\
    \hline
    Freecell & 91.875  & 70.625  \\
    %\hline
    Rovers & 88.125    & 72.5\\
    %\hline
    Trucks & 89    & 73 \\
    %\hline
    Depots & 86.25 & 75.625 \\
    %\hline
    Satellite & 95.5  & 84  \\
    %\hline
    TidyBot & 89    & 80  \\
    
    Blocks World & 89.1  & 74.28 \\
    
    Towers of Hanoi & 84.74 & 74 \\
    
    Barman & 84 & 80 \\
    
    Mystery & 87 & 86 \\
    
    Assembly & 84 & 83 \\
    
    Rockets & 74 & 73 \\
    \hline
    \hline
    \textit{Average} & 86.88 & 77.17 \\
    \hline
    \end{tabular}}
    \caption{Average percentage of applicable preferences that influence decisions for every domain.}
    \label{tab:percentinfluence}
\end{table}

Furthermore, actively acquired preferences influence the decisions in 86.88\% of the cases where the preference applies, compared to 77.17\% for upfront preferences. Table~\ref{tab:percentinfluence} shows the the statistics of every domain. Notice that, while the measures for planning with upfront preferences are almost always less than \textsc{PGPlanner} across all domains, the difference is negligible in the `Mystery' and `Rockets' domains which aligns with what we observe in terms of performance and efficiency. In the `Depots' domain, however, the difference in percentage of decision impacting preference is around $11\%$ (row 4) but the difference in average plan length is substantially high (Figure~\ref{fig:optimality}). On closer inspection we observed that in 2 particular problems in the Depots domain, upfront preferences, though applicable, lead to sub-optimal plans, particularly with substantially high plan length. 

We observe exactly the opposite scenario in the `Towers of Hanoi' domain. Here the difference is around $10\%$, but we observe that difference in performance (average plan length) is not significantly large. This suggests that \textit{we have not compromised on the quality of preferences provided upfront}. Particularly the upfront preference, \textit{``If possible then \textbf{avoid empty pegs}''}, seemed to be effective in influencing the planner towards that part of the search space which mostly resulted in better (shorter) plans. A similar, more prominent case, is the `Freecell' card game domain. The upfront preference \textit{``If possible Then prefer decompositions that allow for immediately finishing a card compared to other choices''} led to shorter plans, even though this preference altered the decisions at only a few points. Clearly, both \textit{Towers of Hanoi} and Freecell are intuitive domains and a little practice allows us to formulate high quality upfront preferences. But other domains are more challenging and it is difficult for an expert to imagine all possible scenarios and formulate useful preferences without knowing the current stage and task. 
%Additionally, among the upfront preferences, only 76.8\% of them alter the planner's decision. %(1.75 among the fired rules, on an average, do not have any impact) in the upfront case.
Also, planning with upfront preferences performs better than random querying in most domains, indicating that the preferences provided were reasonable.  
%This suggests that 
However, \textsc{PGPlanner} is able to elicit more relevant preferences and use them to find more effective plans efficiently. 

One natural question that arises is that the planner assumes that the human preferences are close to optimal (or at least non sub-optimal). This is indeed a correct observation that is true in many human-in-the-loop systems. In systems such as inverse RL~\cite{odom2016active} and probabilistic learning~\cite{odom2016ECML}, the expert's preferences can be explicitly traded-off with trajectories or labeled data respectively. In such cases, the expert's preferences serve to reduce the effect of targeted noise. However, in planning we do not assume access to such trajectories and instead rely on rollout. Thus, an explicit trade-off is not quite sufficient and warrants a deeper investigation which is beyond the scope of the current work. We note that querying a set of different experts based on their expertise level on certain sub-tasks remains an interesting and challenging future direction.

\section{Conclusion}
We present a novel method for preference-guided planning where preferences are actively elicited from human experts. \textsc{PGPlanner} allows for the planner to query only as needed and reduces the burden on the expert to understand the planning process to suggest useful advice. We empirically validate the efficiency and effectiveness of \textsc{PGPlanner} across several domains and demonstrate that it outperforms the baselines even with fewer preferences. %In the future, we aim to extend this work in a few directions. 
Currently, our planner does not validate the preferences, but rather assumes the user is an expert. We will extend to the planner to recommend improvements to the set of preferences. %This allows for a collaborative environment where the planner and user can learn together. Also, 
We will investigate other avenues to obtain preferences including from crowds, transferring across subtasks as well as other domains.

%%%%%%%%%%%%%%%%%%%%%%%%%%%%%%%%%%%%%%%%%%%%%%%%%%%%%%%%%%%%%%%%%%%%%%%%%%%%%%%%%%%%%%%%%%%%%%%%%%%%%%%%%
%% bibliography: see CFP for number of permitted pages

\bibliographystyle{ACM-Reference-Format}  % do not change this line!
\bibliography{biblio}  % put name of your .bib file here

\end{document}